\pdfoutput=1

\documentclass[11pt]{article}

\usepackage[final]{acl}
\usepackage{times}
\usepackage{latexsym}
\usepackage{dsfont} 
\usepackage[T1]{fontenc}
\usepackage[utf8]{inputenc}
\usepackage{float}
\usepackage{microtype}
\usepackage{inconsolata}
\usepackage{graphicx}
\usepackage{amsmath, amssymb, tikz, graphicx}
\usepackage{booktabs}
\usepackage{pifont}
\newcommand{\cmark}{\ding{51}}%
\newcommand{\xmark}{\ding{55}}%

\title{Towards Automated Commentary Generation for Soccer Highlights}

\author{Chidaksh Ravuru \\
  \texttt{chidaksh@cs.unc.edu} \\}

\begin{document}

\maketitle
\section{Abstract}
Automated soccer commentary generation has evolved from template-based systems to advanced neural architectures, aiming to produce real-time descriptions of sports events. While frameworks like SoccerNet-Caption laid foundational work, their inability to achieve fine-grained alignment between video content and commentary remains a significant challenge. Recent efforts such as MatchTime, with its MatchVoice model, address this issue through coarse and fine-grained alignment techniques, achieving improved temporal synchronization. In this paper, we extend MatchVoice to commentary generation for soccer highlights using the GOAL dataset, which emphasizes short clips over entire games. We conduct extensive experiments to reproduce the original MatchTime results and evaluate our setup, highlighting the impact of different training configurations and hardware limitations. Furthermore, we explore the effect of varying window sizes on zero-shot performance. While MatchVoice exhibits promising generalization capabilities, our findings suggest the need for integrating techniques from broader video-language domains to further enhance performance. Our code is available at \href{https://github.com/chidaksh/SoccerCommentary}{https://github.com/chidaksh/SoccerCommentary}.

\section{Related Work}
Automated soccer commentary generation has evolved from early template-based methods to sophisticated neural network approaches. Initial efforts focused on using predefined templates to convert structured match data into natural language summaries. Notable early works include \cite{barzilay-lapata-2005-modeling} which introduced a content selection framework to generate sports summaries from statistics, and \cite{bouayad-agha-etal-2011-content} which improved upon this by emphasizing linguistic variation and content determination. However, such approaches lacked the ability to adapt dynamically to real-time events. 

The development of neural network-based methods marked a significant shift. \cite{van-der-lee-etal-2017-pass} introduced PASS, a data-to-text system for generating tailored soccer reports from match statistics. While this approach improved stylistic adaptability, it remained heavily dependent on predefined rules. \cite{wiseman-etal-2017-challenges} presented a neural model for generating basketball game summaries, highlighting the potential of deep learning for data-driven text generation. However, these efforts primarily focused on generating post-game summaries rather than live commentary.

More recent efforts have focused on large-scale datasets and multimodal learning. SoccerNet-Caption \cite{Mkhallati2023SoccerNetCaption-arxiv} provided 37,000 timestamped commentaries from 471 complete games, enabling training of models capable of generating contextually relevant descriptions. The GOAL dataset \cite{qi2023goalchallengingknowledgegroundedvideo} expanded on this by incorporating external knowledge such as player statistics and strategies, allowing for more accurate and informative descriptions. However, the lack of fine-grained event-commentary alignment remained a limitation.

To address these challenges, MatchTime and its MatchVoice model \cite{rao2024matchtimeautomaticsoccergame} introduced advanced alignment techniques, improving synchronization between commentary and video events. By leveraging multimodal representations and attention mechanisms, MatchTime achieved superior performance in generating real-time descriptions of ongoing events. Additionally, UniSoccer \cite{rao2025universalsoccervideounderstanding} presented the largest multimodal soccer dataset to date, demonstrating the effectiveness of foundation models for soccer-related tasks. Our work builds on these advancements by focusing on commentary generation for highlights using the GOAL dataset, enhancing quality through improved alignment techniques inspired by MatchTime.

\section{Methodology}

The methodology for generating automatic soccer commentary consists of several steps including coarse and fine-grained temporal alignment, as well as commentary generation using the MatchVoice architecture.

\subsection{Addressing Misalignment in SoccerNet-Caption}

The SoccerNet-Caption dataset suffers from significant temporal misalignment between video clips and their corresponding textual commentary. This misalignment arises due to the delayed nature of commentators’ descriptions during live matches, as well as inaccurate timestamp annotations. \cite{rao2024matchtimeautomaticsoccergame} highlighted this issue by analyzing the offset distribution between commentary and visual events, which reveals a substantial proportion of commentaries occurring with significant delays.

\begin{figure}[H]
\centering
\includegraphics[width=1.0\linewidth]{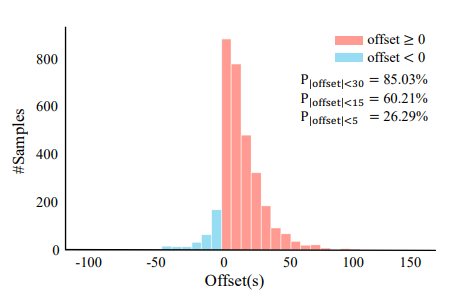}
\caption{Misalignment identified in SoccerNet-Caption dataset where many samples are significantly delayed from the original event.}
\label{fig:misalignment}
\end{figure}

The misalignment issue must be addressed to enhance the performance of models trained on this dataset. To tackle this problem, the MatchTime framework implements a two-level alignment approach: coarse alignment and fine alignment.

\subsection{Coarse Alignment using ASR and LLMs}

The coarse alignment aims to approximately match textual commentary with video content. This is achieved by extracting narration from audio using WhisperX \cite{bain2023whisperxtimeaccuratespeechtranscription}, an automatic speech recognition (ASR) system that produces timestamped text. Since soccer commentary is typically fragmented and colloquial, the transcriptions are then processed by LLaMA-3 \cite{grattafiori2024llama3herdmodels} to generate concise event descriptions for each 10-second video clip. The summarized text is used to infer a rough alignment with the original commentary.

\begin{figure}[H]
\centering
\includegraphics[width=1.0\linewidth]{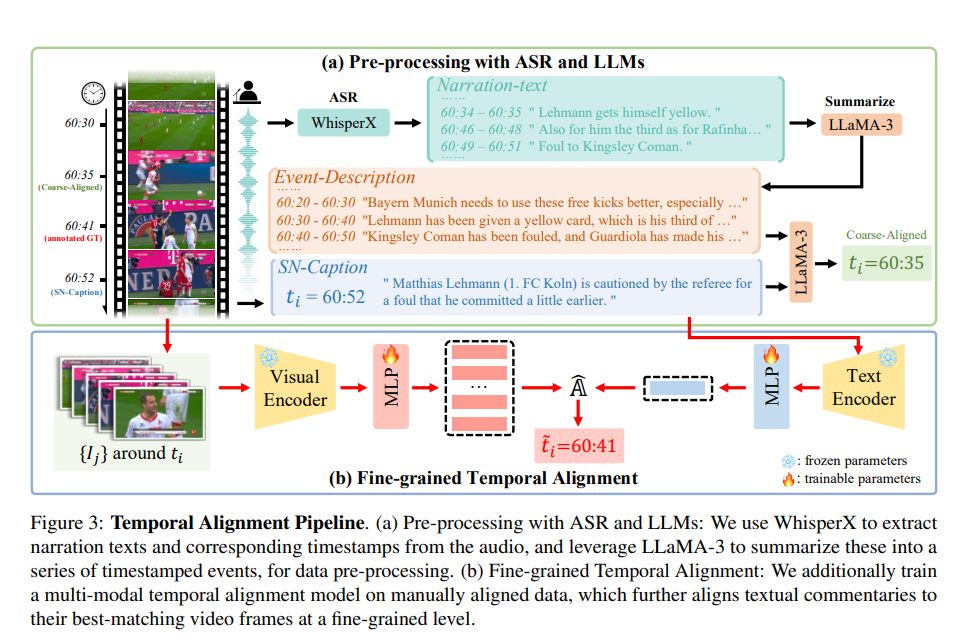}
\caption{MatchTime: Coarse and Fine Alignment Pipeline.}
\label{fig:misalignment}
\end{figure}

\subsection{Fine-grained Temporal Alignment}

The fine-grained alignment process further refines the matching of commentary to visual events through contrastive learning. The alignment model uses a pre-trained CLIP model \cite{radford2021learningtransferablevisualmodels} to encode textual commentaries and video key frames, which are then projected using trainable MLPs $f(\cdot)$ and $g(\cdot)$:

\begin{equation*}
C, V = f(\Phi_{\text{CLIP-T}}(C)), \quad g(\Phi_{\text{CLIP-V}}(V))
\end{equation*}

Where $C \in \mathbb{R}^{k \times d}$ and $V \in \mathbb{R}^{n \times d}$ represent the textual and visual embeddings, respectively.

The affinity matrix $\hat{A} \in \mathbb{R}^{k \times n}$ is computed as:

\begin{equation*}
\hat{A}[i, j] = \frac{C_i \cdot V_j}{\lVert C_i \rVert \lVert V_j \rVert}
\end{equation*}

The alignment model is trained using a contrastive loss function:

\begin{equation*}
\mathcal{L}_{\text{align}} = -\frac{1}{k} \sum_{i=1}^{k} \log \left( \frac{\sum_{j} Y[i, j] \exp(\hat{A}[i, j])}{\sum_{j} \exp(\hat{A}[i, j])} \right)
\end{equation*}

Where $Y \in {0, 1}^{k \times n}$ is the ground-truth label matrix indicating whether commentary $C_i$ corresponds to key frame $V_j$.

\subsection{MatchVoice Architecture}

\begin{figure}[H]
\centering
\includegraphics[width=1.0\linewidth]{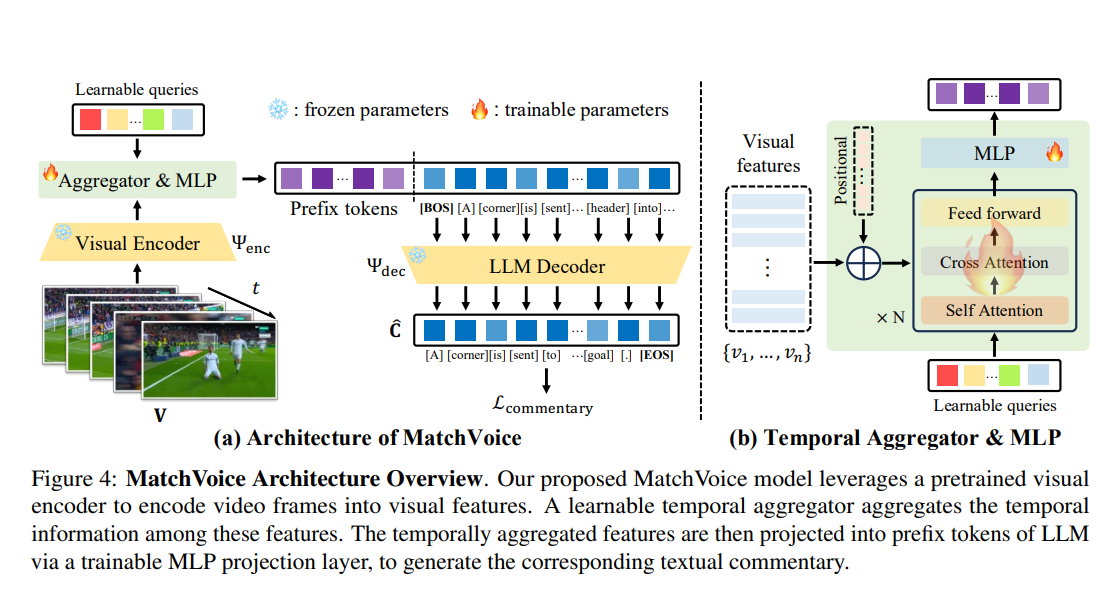}
\caption{MatchVoice Architecture Overview.}
\label{fig:matchvoice_architecture}
\end{figure}

The MatchVoice model is designed to generate natural language commentary from aligned video-text pairs. As illustrated in Figure \ref{fig:matchvoice_architecture}, the model consists of three main components: a frozen pre-trained visual encoder, a Perceiver-like temporal aggregator, and an LLM-based decoder.

\textbf{Visual Encoding.} The frozen visual encoder $\Psi_{\text{enc}}$ processes video frames to obtain framewise features:

\begin{equation*}
{v_1, v_2, \ldots, v_n} = \Psi_{\text{enc}}(V)
\end{equation*}

\textbf{Temporal Aggregation.} The Perceiver-like aggregator $\Psi_{\text{agg}}$ applies transformer decoder layers with learnable queries over visual features to obtain temporally-aware features $F$:

\begin{equation*}
F = \Psi_{\text{agg}}(v_1, v_2, \ldots, v_n)
\end{equation*}

\textbf{Prefix Token Generation.} The aggregated features are mapped to a set of prefix tokens using an MLP projection layer $\Psi_{\text{proj}}$:

\begin{equation*}
\hat{C} = \Psi_{\text{dec}}(\Psi_{\text{proj}}(F))
\end{equation*}

\textbf{Training Objective.} The model is trained to minimize the negative log-likelihood loss:

\begin{equation*}
\mathcal{L}{\text{commentary}} = - \sum_{i} \log P(\hat{C}_i | F)
\end{equation*}

This comprehensive approach ensures that the generated commentary is both temporally accurate and contextually relevant, enhancing the overall quality of the commentary generation process.

\section{Dataset and Evaluation}
To expand the applicability of MatchVoice for commentary generation in short videos such as highlights, we curated the GOAL dataset from YouTube. The dataset is divided into three subsets:

\begin{itemize}
    \item \textbf{Training Data}: 358 videos 
    \item \textbf{Validation Data}: 86 videos 
    \item \textbf{Test Data}: 192 videos 
\end{itemize}

Approximately 500 videos could not be downloaded due to regional restrictions.

\subsection{Evaluation Metrics}
We evaluated the performance of our model using several standard metrics commonly employed in video-language research, including BLEU-1, BLEU-4, METEOR, ROUGE-L, CIDEr, and sBERT. These metrics are consistent with those used in the original MatchTime framework \cite{rao2024matchtimeautomaticsoccergame}. 

\subsection{Hardware Configuration}
All experiments were conducted using a single NVIDIA L-40 GPU on the Longleaf computing cluster, ensuring consistent and efficient training across various configurations.

\section{Experiments}
\label{sec:experiments}

In this section, we present the experimental results for MatchVoice trained on SoccerNet-Caption and MatchTime datasets, along with our reproduced results and comparison. Additionally, we evaluate the impact of different window sizes on the performance of MatchVoice for the GOAL dataset.

\subsection{Comparison of Original and Reproduced Results}
We compare the results reported in the original paper with our reproduced results using the same dataset and evaluation metrics. The original results are summarized in Table \ref{tab:original_results}, while our reproduced results are presented in Table \ref{tab:reproduced_results}. The absolute differences between these results are shown in Table \ref{tab:difference_results}. 

\begin{table*}[htbp]
\centering
\begin{tabular}{|l|c|c|c|c|c|c|}
\hline
\textbf{Features} & \textbf{BLEU-1} & \textbf{BLEU-4} & \textbf{METEOR} & \textbf{ROUGE-L} & \textbf{CIDER} & \textbf{GPT-score} \\ \hline
Baidu Aligned & \textbf{\textcolor{red}{31.42}} & \textbf{\textcolor{red}{8.92}} & \textbf{\textcolor{red}{26.12}} & \textbf{\textcolor{red}{29.66}} & \textbf{\textcolor{red}{38.42}} & \textbf{\textcolor{red}{7.08}} \\ \hline
Baidu Misaligned & \textcolor{blue}{30.32} & \textcolor{blue}{8.45} & \textcolor{blue}{25.25} & 29.40 & \textcolor{blue}{33.84} & \textcolor{blue}{7.07} \\ \hline
CLIP Aligned & 29.56 & 6.90 & {24.62} & \textcolor{blue}{31.25} & {28.66} & 6.82 \\ \hline
CLIP Misaligned & 28.65 & 6.62 & 24.20 & 27.33 & 24.35 & 6.78 \\ \hline
\end{tabular}
\caption{Original Results Reported in the Paper - Aligned vs. Misaligned Models. \textcolor{red}{Red} denotes the best performance and the second-best performance in \textcolor{blue}{Blue}.}
\label{tab:original_results}
\end{table*}

\begin{table*}[htbp]
\centering

\begin{tabular}{|l|c|c|c|c|c|c|}
\hline
\textbf{Features} & \textbf{BLEU-1} & \textbf{BLEU-4} & \textbf{METEOR} & \textbf{ROUGE-L} & \textbf{CIDER} & \textbf{sBERT} \\ \hline
Baidu Aligned & \textbf{\textcolor{red}{29.643}} & \textbf{\textcolor{red}{8.227}} & \textbf{\textcolor{red}{26.491}} & \textbf{\textcolor{red}{26.316}} & \textbf{\textcolor{red}{34.452}} & \textbf{\textcolor{red}{68.981}} \\ \hline
Baidu Misaligned & {27.329} & \textcolor{blue}{6.703} & 23.494 & 22.765 & \textcolor{blue}{26.511} & 63.341 \\ \hline
CLIP Aligned & \textcolor{blue}{27.433} & 6.219 & \textcolor{blue}{25.255} & \textcolor{blue}{24.456} & 24.965 & \textcolor{blue}{66.464} \\ \hline
CLIP Misaligned & 24.625 & 4.250 & 22.701 & 20.623 & 15.053 & 60.882 \\ \hline
\end{tabular}
\caption{Reproduced Results. \textcolor{red}{Red} denotes the best performance and the second-best performance in \textcolor{blue}{Blue}.}
\label{tab:reproduced_results}
\end{table*}

\begin{table*}[htbp]
\centering
\begin{tabular}{|l|c|c|c|c|c|c|}
\hline
\textbf{Features} & \textbf{BLEU-1} & \textbf{BLEU-4} & \textbf{METEOR} & \textbf{ROUGE-L} & \textbf{CIDER} & \textbf{sBERT} \\ \hline
Baidu Aligned & 0.677 & 0.223 & -1.241 & 3.084 & -0.612 & -61.911 \\ \hline
Baidu Misaligned & 2.281 & 0.127 & 1.886 & 2.515 & -5.901 & -56.621 \\ \hline
CLIP Aligned & 1.217 & 0.561 & -0.905 & 2.874 & 2.765 & -59.684 \\ \hline
CLIP Misaligned & -0.005 & 0.000 & -0.001 & -0.003 & 0.003 & -0.802 \\ \hline
\end{tabular}
\caption{Difference Between Original and Reproduced Results (Original - Reproduced).}
\label{tab:difference_results}
\end{table*}

\begin{table*}[htbp]
\centering
\begin{tabular}{|l|c|c|c|c|c|c|}
\hline
\textbf{Model} & \textbf{BLEU-1} & \textbf{BLEU-4} & \textbf{METEOR} & \textbf{ROUGE-L} & \textbf{CIDER} & \textbf{sBERT} \\ \hline
GOAL Test (Window = 3) & 6.832 & 0.056 & 14.932 & 8.235 & 1.079 & 33.736 \\ \hline
GOAL Test (Window = 5) & 6.905 & \textcolor{red}{0.107} & 14.889 & 8.312 & 1.130 & 33.868 \\ \hline
GOAL Test (Window = 10) & \textbf{\textcolor{red}{7.120}} & \textcolor{blue}{0.095} & \textbf{\textcolor{red}{15.157}} & \textcolor{blue}{8.449} & \textbf{\textcolor{red}{1.264}} & \textbf{\textcolor{red}{34.299}} \\ \hline
GOAL Test (Window = 15) & \textcolor{blue}{7.024} & 0.091 & \textcolor{blue}{15.105} & \textcolor{red}{8.493} & \textcolor{blue}{1.232} & \textcolor{blue}{34.010} \\ \hline
\end{tabular}
\caption{Window Size Comparison for MatchVoice (Zero-shot on GOAL Dataset). \textcolor{red}{Red} denotes the best performance and the second-best performance in \textcolor{blue}{Blue}.}
\label{tab:goal_window_comparison}
\end{table*}

\begin{table*}[htbp]
\centering
\begin{tabular}{|p{4.5cm}|c|c|c|c|c|c|}
\hline
\textbf{Model} & \textbf{BLEU-1} & \textbf{BLEU-4} & \textbf{METEOR} & \textbf{ROUGE-L} & \textbf{CIDER} & \textbf{sBERT} \\ \hline
SN-Caption-test-align before Fine Tuning & \textcolor{red}{27.433} & \textcolor{red}{6.219} & \textcolor{red}{25.255} & \textcolor{red}{24.456} & \textcolor{red}{24.965} & \textcolor{red}{66.464} \\ \hline
SN-Caption-test-align after Fine Tuning & 20.12 & 4.75 & 23.89 & 20.35 & 21.23 & 63.421 \\ \hline
GOAL Test before Fine Tuning & {7.120} & 0.095 & 15.157 & {8.449} & 1.264 & 34.299 \\ \hline
GOAL Test after Fine Tuning & \textcolor{blue}{8.24} & \textcolor{blue}{0.401} & \textcolor{blue}{16.138} & \textcolor{blue}{8.92} & \textcolor{blue}{2.18} & \textcolor{blue}{38.76} \\ \hline

\end{tabular}
\caption{Comparison of MatchVoice Performance Before and After Fine-Tuning. The highest value for each metric is highlighted in \textcolor{red}{Red}, while the second-highest is highlighted in \textcolor{blue}{Blue}.}
\label{tab:fine_tuning_comparison}
\end{table*}

\subsection{Discussion}
The difference in the reproduced results compared to the original paper can be attributed to several factors. Firstly, the original paper utilized A100 GPUs which have superior memory management and higher computation power compared to the L40 GPU used in our experiments. This hardware limitation likely affected the training stability and efficiency of the model, especially during fine-tuning. Additionally, subtle differences for example, (\cite{rao2024matchtimeautomaticsoccergame} used a batchsize of 64 with 16 workers, but due to resource constraints, we used a batch size of 32 with 8 workers) could have also contributed to the observed discrepancies.

\subsection{Window Size Experiments}
We also experimented with varying window sizes for MatchVoice in a zero-shot setting on the GOAL dataset. The results are presented in Table \ref{tab:goal_window_comparison}. Based on the experiments, we decided to use a window-size of 10 for rest of the experiments on GOAL dataset.

\subsection{Fine-Tuning on GOAL}
The results presented in Table \ref{tab:fine_tuning_comparison} demonstrate the effectiveness of fine-tuning MatchVoice on the GOAL dataset. Despite being trained on only 100 videos due to resource constraints, the model shows noticeable improvements across all evaluation metrics for the GOAL test set. Specifically, BLEU-1 and BLEU-4 scores improved significantly from 7.120 to 8.24 and from 0.095 to 0.401, respectively. METEOR, ROUGE-L, CIDER, and sBERT scores also showed considerable improvements, indicating that the model has effectively learned from the fine-tuning process.

Importantly, the metrics for the SN-Caption-test-align dataset did not experience a substantial decrease after fine-tuning, particularly with respect to METEOR and sBERT. This observation suggests that the model is successfully retaining its ability to capture semantic similarity and paraphrasing, which is essential for commentary generation. The relative stability of these metrics underscores the model's capacity to generalize across different datasets, even with limited fine-tuning resources.

\subsection{Ablation Study on Fine-tuning}

To understand the contribution of different modalities in the fine-tuning process, we conducted an ablation study by selectively enabling vision and language components during training. The results are presented in Table~\ref{tab:ablation}.

\begin{table*}[htbp]
\centering
\begin{tabular}{|c|c|c|c|c|c|c|c|}
\hline
\textbf{Vision} & \textbf{Language} & \textbf{BLEU-1} & \textbf{BLEU-4} & \textbf{METEOR} & \textbf{ROUGE-L} & \textbf{CIDEr} & \textbf{sBERT} \\ \hline
\xmark & \xmark & 7.125 & 0.105 & 14.889 & 8.462 & 1.469 & 33.373 \\ \hline
\cmark & \xmark & 7.491 & 0.309 & 15.300 & 8.691 & 1.840 & 34.170 \\ \hline
\xmark & \cmark & 8.102 & 0.401 & 15.780 & 8.776 & 2.050 & 36.896 \\ \hline
\cmark & \cmark & \textbf{8.240} & \textbf{0.382} & \textbf{16.138} & \textbf{8.920} & \textbf{2.180} & \textbf{38.760} \\ \hline

\end{tabular}
\caption{Ablation study on finetuning. Checkmarks indicate whether the vision encoder and/or the language model were fine-tuned.}
\label{tab:ablation}
\end{table*}

From the table, we observe that:

\begin{itemize}
    \item Fine-tuning only the vision encoder (row 2) yields improvements across all metrics compared to the frozen baseline (row 1), especially in CIDEr (+0.37) and BLEU-4 (+0.204).
    \item Fine-tuning only the language model (row 3) leads to even more pronounced gains in BLEU, METEOR, and semantic similarity (sBERT).
    \item Joint fine-tuning of both the vision encoder and the language model (row 4) produces the best overall performance, confirming that full end-to-end optimization leads to richer visual grounding and better linguistic fluency.
\end{itemize}

This ablation underscores the importance of synergistic tuning of both visual and linguistic components in multimodal models like ours.

\section{Zero-shot Inference}

To evaluate baseline generalization without task-specific supervision, we conducted zero-shot inference using two pretrained video-language models: \textbf{Video-ChatGPT} and \textbf{UniSoccer/MatchVision}. These models were tested on short 10–15 second video clips centered around events in our Goal dataset, created by cropping longer videos to satisfy the input constraints of the models (typically less than 30 seconds). This ensured compatibility and focused event relevance.

\subsection{Video-ChatGPT Zero-shot Results}
\begin{figure*}[htbp]
    \centering
    \includegraphics[width=0.85\textwidth]{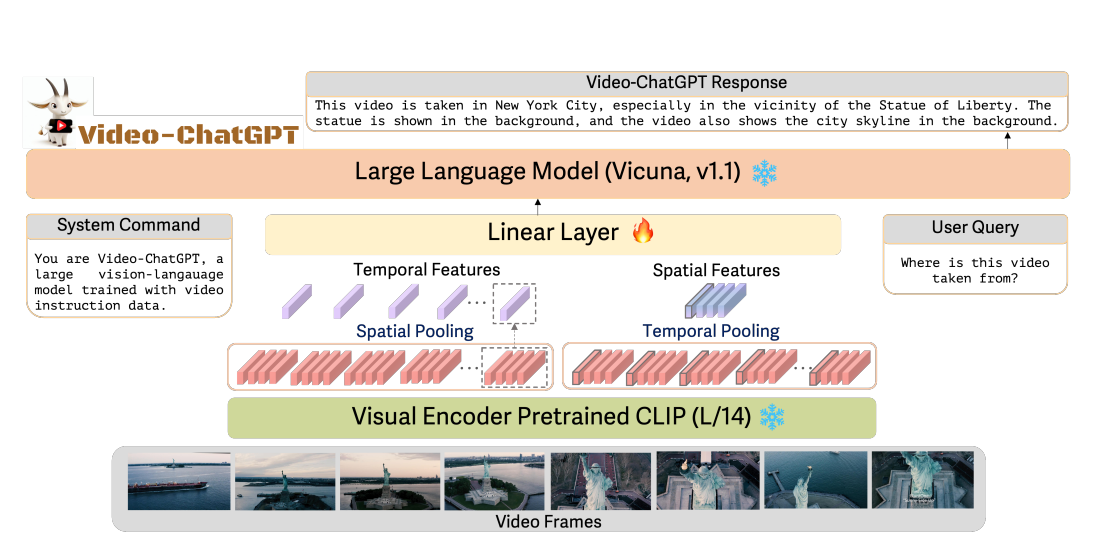}
    \caption{Architecture of Video-ChatGPT}
    \label{fig:videogpt}
\end{figure*}

We first evaluated the Video-ChatGPT \cite{maaz2024videochatgptdetailedvideounderstanding} model, which follows a CLIP-based visual encoder and Vicuna-based LLM decoder architecture (Figure~\ref{fig:videogpt}). As the model was primarily trained for general video summarization and lacks domain-specific grounding in soccer, its raw zero-shot performance was low.

The initial results, without any domain adaptation, are summarized below:

\begin{table*}[h]
\centering
\begin{tabular}{lccccc}
\toprule
\textbf{Setting} & \textbf{BLEU-1} & \textbf{BLEU-4} & \textbf{ROUGE-L} & \textbf{CIDER} & \textbf{sBERT} \\
\midrule
Zero-Shot & 1.484 & 0.017 & 4.166 & 0.005 & 28.380 \\
Zero-Shot + Post-Processing & 2.260 & 0.062 & 5.935 & 0.214 & 31.842 \\
\bottomrule
\end{tabular}
\caption{Video-ChatGPT zero-shot results before and after post-processing.}
\label{tab:video-gpt}
\end{table*}

Post-processing involved anonymizing the generated commentaries to match the Goal dataset style by replacing names such as \texttt{player} and \texttt{team} with tokens like \texttt{[PLAYER]}, \texttt{[TEAM]}, etc., and removing subjective or filler language like ``I am asked to...'' or ``Let me describe...''. This led to noticeable improvements across all metrics. However, performance remained limited due to distribution shift: the model was not trained on soccer-specific language and lacked grounding in event terminology.

\subsection{UniSoccer / MatchVision Zero-shot Results}

\begin{figure*}[htbp]
    \centering
    \includegraphics[width=0.92\textwidth]{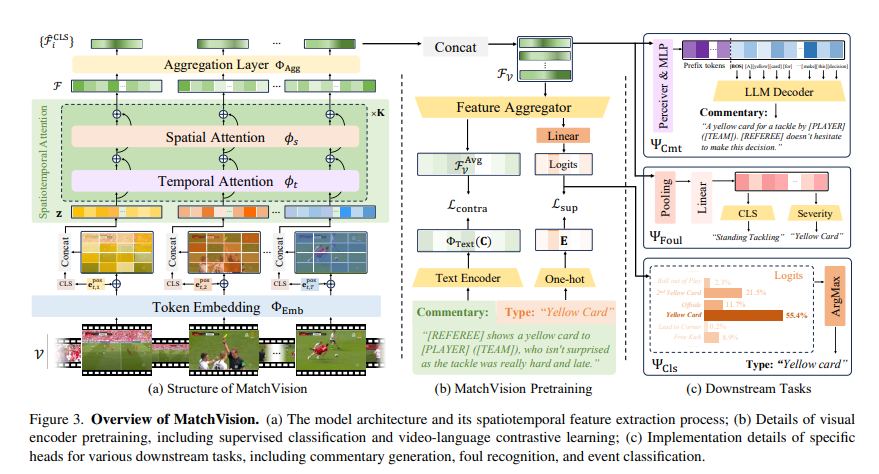}
    \caption{Architecture of MatchVision (UniSoccer)}
    \label{fig:unisoccer}
\end{figure*}

We also evaluated the UniSoccer \cite{rao2025universalsoccervideounderstanding} model, which integrates spatiotemporal attention over video frames and supports downstream tasks like commentary generation, foul classification, and event recognition (Figure~\ref{fig:unisoccer}). Despite its specialization in soccer events, fine-tuning on the Goal dataset was infeasible due to lack of explicit event labels. Attempts at automatic weak-labeling (e.g., detecting ``goal'' or ``corner'' from commentaries) were noisy and unreliable due to short or vague annotations.

The zero-shot and fine-tuned performance of UniSoccer on the SN-Caption-test-align subset is as follows:

\begin{table*}[h]
\centering
\begin{tabular}{lccccc}
\toprule
\textbf{Model} & \textbf{BLEU-1} & \textbf{BLEU-4} & \textbf{ROUGE-L} & \textbf{CIDER} & \textbf{sBERT} \\
\midrule
UniSoccer (Zero-shot) & 21.65 & 3.27 & 21.02 & 17.79 & 12.90 \\
UniSoccer (Fine-tuned on MatchTime) & 27.49 & 6.96 & 24.50 & 23.33 & 30.81 \\
\bottomrule
\end{tabular}
\caption{UniSoccer performance on SN-Caption-test-align before and after fine-tuning.}
\label{tab:unisoccer}
\end{table*}

Although fine-tuning on MatchTime improved scores, the results were still weaker than those of MatchVoice, which had direct supervision from the same dataset. We attribute this gap to dataset distribution shift: MatchVision was trained on SoccerNet-v2, a broader but stylistically different dataset from MatchTime.

\begin{figure*}[htbp]
    \centering
    \includegraphics[width=0.95\textwidth]{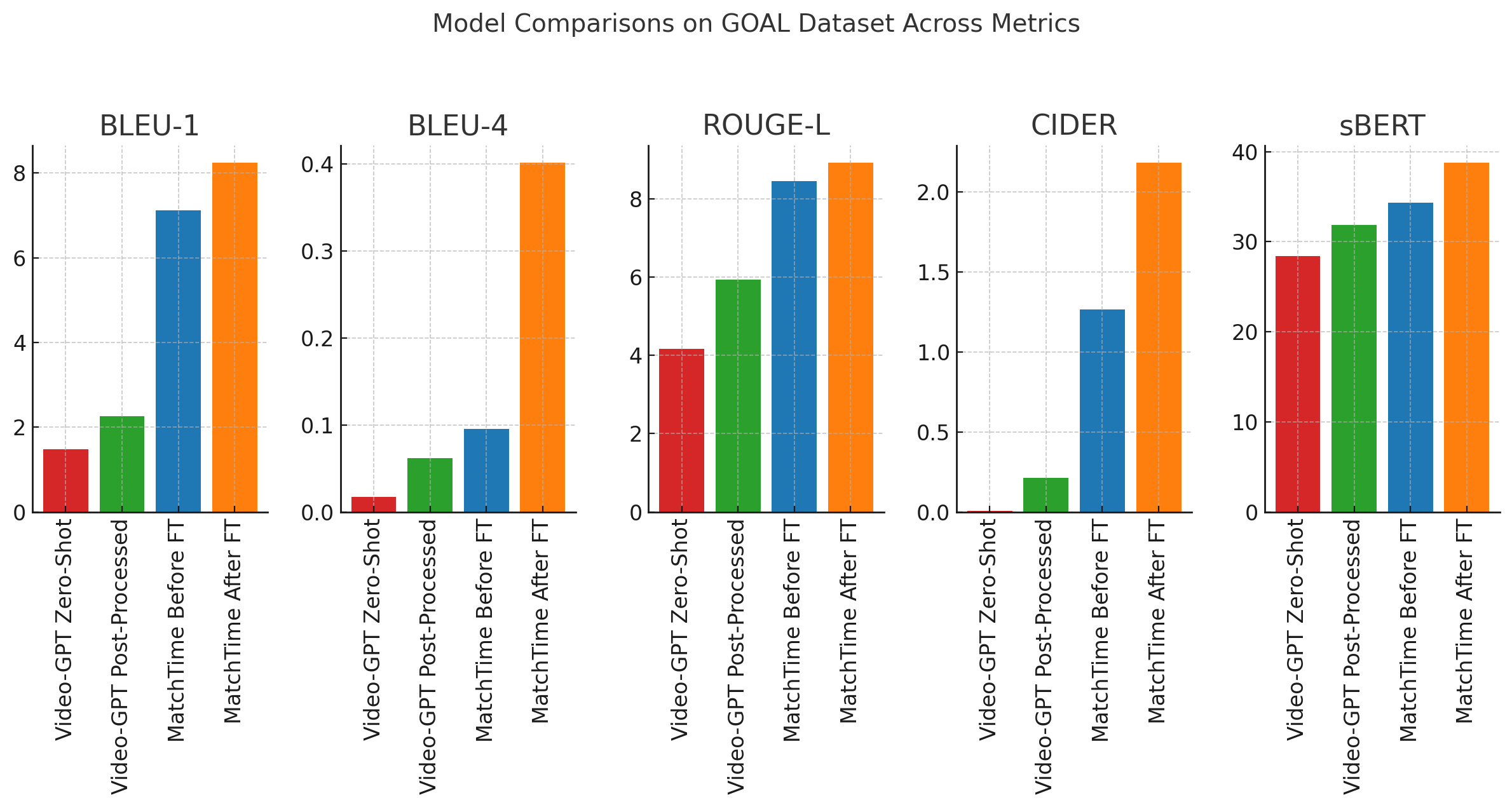}
    \caption{
        \textbf{Comparative Evaluation of Video-GPT and MatchTime on the GOAL Dataset.}
        The figure reports performance across five metrics: BLEU-1, BLEU-4, ROUGE-L, CIDER, and sBERT.
    }
    \label{fig:goal-metrics-comparison}
\end{figure*}

The performaces of all the models is summarized in Figure \ref{fig:goal-metrics-comparison}. Post-processing (token anonymization and style cleaning) improves Video-GPT’s zero-shot performance. Fine-tuning MatchTime on the GOAL dataset leads to the best results across all metrics, demonstrating the benefit of supervised domain alignment. Interestingly, MatchTime Before FT already outperforms Video-GPT post-processed, indicating a stronger inductive bias from soccer-specific pretraining.

\section{Potential Issues in general Soccer Commentary Frameworks}

\begin{itemize}
    \item \textbf{Ground Truth Alignment:} One of the key challenges in soccer commentary generation is the misalignment between ground truth captions and actual human commentary. Future work should explore techniques for aligning temporally relevant commentary with specific game events.

    \item \textbf{Entity Identification and De-Anonymization:} The current datasets use anonymized tokens like \texttt{[PLAYER]}, \texttt{[TEAM]}, etc., which hinders the generation of personalized and context-rich commentary. A potential direction is the development of models capable of entity resolution and dynamic name generation.

    \item \textbf{Expanding Dataset Scale and Diversity:} To improve model generalization and robustness, larger and more diverse datasets are required. Incorporating extensive datasets such as \textit{SoccerReplay-1988 \cite{rao2025universalsoccervideounderstanding}} could enable better training for event-rich, multi-view, and multilingual soccer scenarios.
\end{itemize}

\newpage
\bibliography{custom}
\end{document}